%% file: main.tex
\documentclass[conference]{IEEEtran}
\IEEEoverridecommandlockouts
\usepackage[square,sort&compress,comma,numbers]{natbib}

\usepackage{times}
\usepackage{epsfig}
\usepackage{graphicx}
\usepackage{amsmath}
\usepackage{amssymb}
\usepackage{subfig}

\usepackage{booktabs}
\usepackage[table]{xcolor}
\usepackage{enumitem}

\usepackage{pifont}

\usepackage[ruled,noend]{algorithm2e}

\SetCommentSty{mycommfont}
\SetKwInput{KwInput}{Input}
\SetKwInput{KwOutput}{Output}

\definecolor{brickred}{rgb}{0.8, 0.25, 0.33}
\definecolor{brickred2}{rgb}{0.25, 0.8, 0.33}
\newcommand{\xm}{\color{brickred2}{\ding{51}}}%
\newcommand{\cm}{\color{brickred}{\ding{55}}}%

\usepackage[colorlinks=true,
citecolor=blue,
filecolor=black,
linkcolor=blue,
urlcolor=blue]{hyperref}

\input notation.tex

\input macros.tex

\usepackage{enumitem} 

\newcommand{\OURS}{\textsc{ZeroQ}\xspace}

\newcommand{\rg}{Distilled Data\xspace}

\begin{document}
\title{ZeroQ: A Novel Zero Shot Quantization Framework}

\author{
Yaohui Cai$^{*,1}$\thanks{$^{*}$Equal contribution.}, Zhewei Yao$^{*,2}$, Zhen Dong,$^{*,2}$ \\
Amir Gholami$^2$, Michael W. Mahoney$^2$, Kurt Keutzer$^2$\\
$^1$Peking University; \ $^2$University of California, Berkeley\\
{\small \{zheweiy, zhendong, amirgh, mahoneymw, and keutzer\}@berkeley.edu
\ caiyaohui@pku.edu.cn}

}
\maketitle
\input {_s0_abstract.tex}
\input {_s1_intro.tex}
\input {_s2_related_work.tex}

\input {_s3_methods.tex}
\input {_s4_results.tex}

\input {_s5_ablation.tex}

\input {_s6_conclusions.tex}

{\small
\bibliographystyle{ieee_fullname}
\bibliography{ref}
}
\clearpage
\onecolumn
\appendix
\input {_s7_appendix.tex}

\end{document}

%% file: notation.tex
\newcommand{\M}{\mathcal M}

\def\tt{{\bf t}}

\def\0{{\bf 0}}
\def\1{{\bf 1}}

\def\M{\mathcal M}

\def\L{\mathcal L}

%% file: macros.tex
\newcommand\aref{Algorithm~\ref}

\newcommand\eref{Eq.~\ref}
\newcommand\fref{Figure~\ref}
\newcommand\tref{Table~\ref}
\newcommand\sref{Section~\ref}

\newcommand\ha{ \rowcolor{orange!0}}

\newcommand\hc{ \rowcolor{orange!40}}

\newcolumntype{g}{>{\columncolor{orange!40}}c}

\def\0{{\bf 0}}

\def\R{{\mathbb R}}

\newcommand{\AV}{\xspace{MP}}

%% file: _s0_abstract.tex
\begin{abstract}
Quantization is a promising approach for reducing the inference time and memory footprint of neural networks.
However, most existing quantization methods require access to the original training dataset for retraining during quantization.
This is often not possible for applications with sensitive or proprietary data, e.g., due to privacy and security concerns.
Existing zero-shot quantization methods use different heuristics to address this, but they result in poor performance, especially when quantizing to ultra-low precision.
Here, we propose \OURS, a novel zero-shot quantization framework to address this.
\OURS enables mixed-precision quantization without any access to the training or validation data. 
This is achieved by optimizing for a Distilled Dataset, which is engineered to match the statistics of batch normalization across different layers of the network.
\OURS supports both uniform and mixed-precision quantization. 
For the latter, we introduce a novel Pareto frontier based method to automatically 
determine the mixed-precision bit setting for all layers, with no manual search involved.
We extensively test our proposed method on a diverse set of models, including ResNet18/50/152, MobileNetV2, ShuffleNet, SqueezeNext, and InceptionV3 on ImageNet, as well as RetinaNet-ResNet50 on the Microsoft COCO dataset.
In particular, we show that \OURS can achieve 1.71\% higher accuracy on MobileNetV2, as compared to the recently proposed DFQ~\cite{nagel2019data} method.
Importantly, \OURS has a very low computational overhead, and it can finish the entire quantization process in less than 30s (0.5\% of one epoch training time of ResNet50 on ImageNet).
We have open-sourced the \OURS framework\footnote{https://github.com/amirgholami/ZeroQ}.
\end{abstract}

%% file: _s1_intro.tex
\section{Introduction}
\label{sec:intro}

Despite the great success of deep Neural Network (NN) models in various domains, the deployment of modern NN models  at the edge has been challenging due to their prohibitive memory
footprint, inference time, and/or energy consumption.
With the current hardware support for low-precision computations,  quantization has become a popular procedure to address these challenges. 
By quantizing the floating point values of weights and/or activations in a NN to integers, the model size can be shrunk significantly, without any modification to the architecture. 
This also allows one to use reduced-precision Arithmetic Logic Units (ALUs) which
are faster and more power-efficient, as compared to floating point ALUs. More importantly, quantization
reduces memory traffic volume, which is a significant source of energy consumption~\cite{horowitz20141}.

\begin{figure*}[t]
\centering
\includegraphics[width=.95\textwidth]{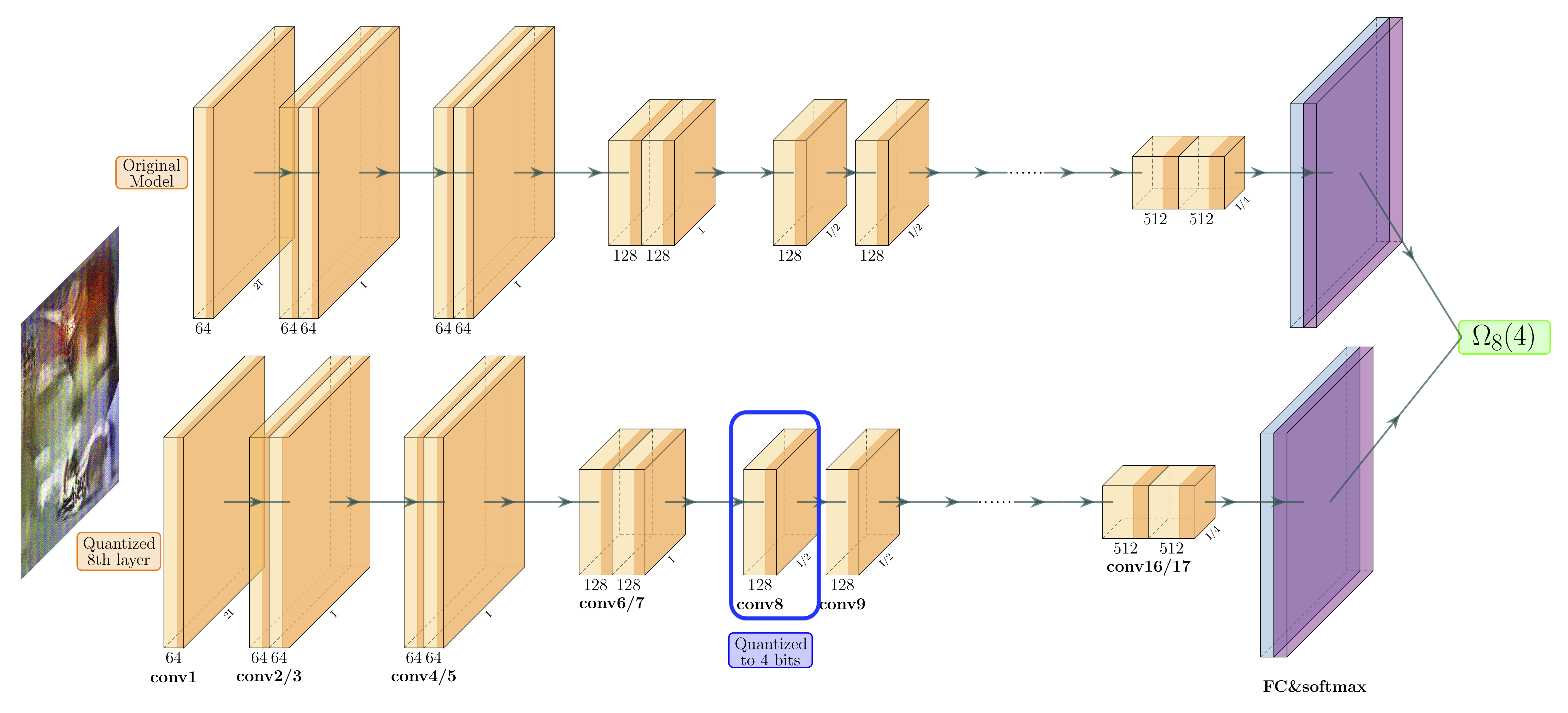}
\caption{
Illustration of sensitivity computation for ResNet18 on ImageNet. 
The figure shows how we compute the sensitivity of the 8-th layer when quantized to 4-bit ($\Omega_8(4)$) according to \eref{eq:sensivity_formula}.
We feed \rg into the full-precision ResNet18 (top), and the same model except quantizing the 8-th layer to 4-bit (bottom) receptively. 
The sensitivity of the 8-th layer when quantized to 4-bit $\Omega_8(4)$ is defined as the KL-divergence between the output of these two models.
For simplicity, we omit the residual connections here, although the same analysis is applied to the residual connections in \OURS.
}
  \label{fig:resnet18_sensitivity_illustration}
\end{figure*}

However, quantizing a model from single precision to low-precision
often results in significant accuracy degradation.
One way to alleviate this is to perform the so-called
quantization-aware fine-tuning~\cite{zhou2016dorefa, rastegari2016xnor, choi2018pact, Zhang_2018_ECCV, zhou2017incremental, jacob2018quantization}
to reduce the performance gap
between the original model and the quantized model. Basically, this is a retraining procedure that
is performed for a few epochs to adjust the NN parameters to reduce accuracy drop.
However, quantization-aware fine-tuning can be computationally expensive and time-consuming.
For example, in online learning situations, where a model needs to be constantly updated on new data and deployed every few hours, there may not be enough time for the fine-tuning procedure to finish.
More importantly, in many real-world scenarios, the training dataset is sensitive or proprietary, meaning that it is not
possible to access the dataset  
that was used to train the model.
Good examples are medical data, bio-metric data, or user data used in recommendation systems.

To address this, recent work has proposed post-training quantization~\cite{Kravchik_2019_ICCV, nagel2019data, zhao2019improving, banner2018post}, which directly quantizes NN models without fine-tuning.
However, as mentioned above, these methods result in non-trivial performance degradation, especially for low-precision quantization.
Furthermore, previous post-training quantization methods usually require limited (unlabeled) data to assist the post-training quantization. 
However, for cases such as MLaaS (e.g., Amazon AWS and Google Cloud), it may not be possible to access
any of the training data from users.
An example application case is health care information which cannot be uploaded 
to the cloud due to various privacy issues and/or regulatory constraints.
Another shortcoming is that often post-quantization methods~\cite{meller2019same,zhao2019improving,banner2018post} only focus on standard NNs such as ResNet~\cite{he2016deep} and InceptionV3~\cite{szegedy2016rethinking}
for image classification, and they do not consider more demanding tasks such as object detection.

In this work, we propose \OURS, a novel zero-shot quantization scheme to overcome the issues mentioned above.
In particular, \OURS allows quantization of NN models, without any access to any training/validation data.
It uses a novel approach to automatically compute  a mixed-precision configuration without any expensive search. 
In particular, our contributions are as~follows.

\begin{itemize}[noitemsep,topsep=0pt,parsep=0pt,partopsep=0pt,leftmargin=*]
    \item 
    We propose an optimization formulation to generate 
    \rg, i.e., synthetic data  engineered to match the
    statistics of batch normalization layers.
    This reconstruction has a small computational overhead. For example, it
    only takes 3s (0.05\% of one epoch training time) to generate 32 images for ResNet50 on ImageNet on an 8-V100 system. 
    \item 
    We use the above reconstruction framework to perform sensitivity analysis between the quantized and the original model. 
    We show that the \rg matches the sensitivity of the original training data (see Figure \ref{fig:resnet18_sensitivity_illustration} and~\tref{tab:ablation_recon} for details).
    We then use the \rg, instead of original/real data, to perform post-training quantization.
    The entire sensitivity computation here only costs 12s (0.2\% of one epoch training time) in total for ResNet50. 
    Importantly, we never use any training/validation data for the entire process.
    \item Our framework supports both uniform and mixed-precision quantization.
    For the latter,
    we propose a novel automatic precision selection method based on a Pareto frontier optimization (see Figure \ref{fig:pareto_frontier} for illustration).
    This is achieved by computing the quantization sensitivity based on the \rg with small
    computational overhead. For example, we are able to determine automatically the mixed-precision setting in under 14s for ResNet50.
\end{itemize}
We extensively test our proposed \OURS framework on a wide range of NNs for image classification and object detection tasks,  achieving state-of-the-art quantization results in all tests. 
In particular, we present quantization results for both standard models (e.g., ResNet18/50/152 and InceptionV3) and efficient/compact models (e.g., MobileNetV2, ShuffleNet, and SqueezeNext) for image classification task. 
Importantly, we also test \OURS for object detection on Microsoft COCO dataset~\cite{lin2014microsoft} with RetinaNet~\cite{lin2017focal}. 
Among other things, we show that \OURS achieves 1.71\% higher accuracy on MobileNetV2 as compared to the recently proposed DFQ~\cite{nagel2019data} method.

%% file: _s2_related_work.tex
\section{Related work}
\begin{figure*}[!htbp]
\centering
\includegraphics[width=.45\textwidth]{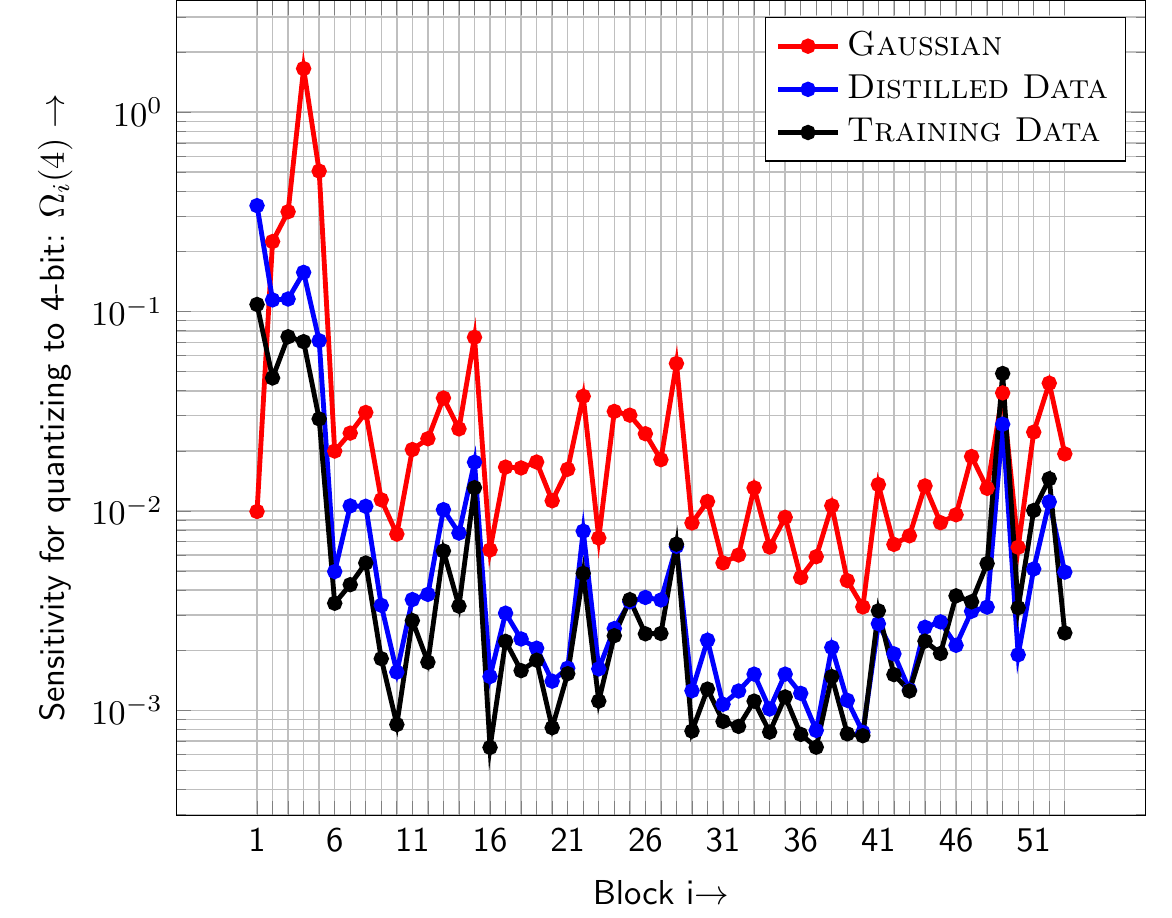}
\includegraphics[width=.45\textwidth]{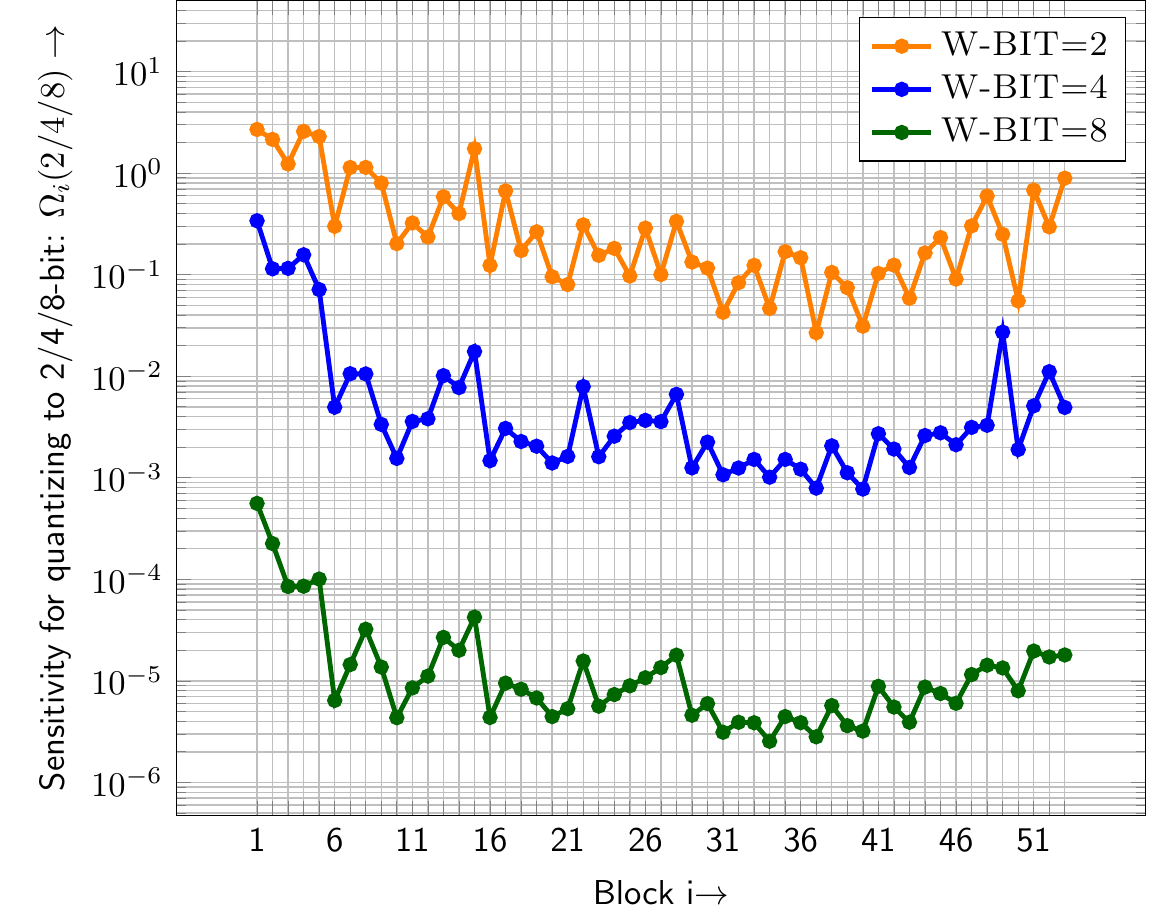}
\caption{
  (Left) Sensitivity of each layer in ResNet50 when quantized to 4-bit weights, measured with different kinds of data (red for Gaussian, blue for \rg, and black for training data). 
  (Right) Sensitivity of ResNet50 when quantized to 2/4/8-bit weight precision (measured with \rg). 
}
  \label{fig:sensitivity_resnet50_mobilenetv2}
\end{figure*}

Here we provide a brief (and by no means extensive) review of the related work in literature.
There is a wide range of methods besides quantization which have been proposed to
address the prohibitive 
memory footprint and inference latency/power of modern NN architectures.
These methods are typically orthogonal to quantization, and they include efficient neural architecture design~\cite{iandola2016squeezenet,gholami2018squeezenext,howard2017mobilenets,sandler2018mobilenetv2, zhang2018shufflenet}, knowledge distillation~\cite{hinton2015distilling,romero2014fitnets}, 
model pruning~\cite{han2015learning, mao2017exploring, li2016pruning}, and hardware and NN co-design~\cite{gholami2018squeezenext,kwon2018co}.
Here we focus on quantization~\cite{asanovic1991experimental,courbariaux2015binaryconnect,rastegari2016xnor,wu2016quantized,li2016ternary,zhu2016trained,zhou2017incremental,zhou2016dorefa,choi2018pact, dong2019hawq, Zhang_2018_ECCV},
which compresses the model by reducing the bit precision used to represent
parameters and/or activations.
An important challenge with quantization is that it can lead to significant performance degradation, especially in ultra-low bit precision settings.
To address this, existing methods propose quantization-aware fine-tuning to recover lost performance~\cite{krishnamoorthi2018quantizing, jacob2018quantization, baskin2019cat}.
Importantly, this requires access to the full dataset that was used to train the original model. Not only can this be very time-consuming, but often access to training
data is not possible.

To address this, several papers focused on developing post-training quantization methods (also referred to as post-quantization), without any fine-tuning/training. 
In particular, \cite{Kravchik_2019_ICCV} proposes the OMSE method to optimize the $L_2$ distance between the quantized tensor and the original tensor. Moreover,
\cite{banner2018post} proposed the so-called ACIQ method to analytically compute the clipping range, as well as the per-channel bit allocation for NNs, and it achieves relatively good testing performance. 
However, they use per-channel quantization for activations,
which is difficult for efficient hardware implementation in practice.
In addition, \cite{zhao2019improving} proposes an outlier channel splitting (OCS) method to solve the outlier channel problem. 
However, these methods require access to limited data to reduce the performance
drop~\cite{Kravchik_2019_ICCV,banner2018post,zhao2019improving,meller2019same,lee2018quantization}.

The recent work of~\cite{nagel2019data} proposed Data Free Quantization (DFQ). It further pushes post-quantization to zero-shot scenarios, where neither training nor testing data are accessible during quantization. The work of~\cite{nagel2019data} uses a weight equalization scheme~\cite{meller2019same} to remove outliers in both weights and activations, and they achieve similar results with layer-wise quantization, as compared to previous post-quantization work with channel-wise quantization~\cite{krishnamoorthi2018quantizing}. However,~\cite{nagel2019data} their performance significantly degrades when NNs are quantized to 6-bit or lower.

A recent concurrent paper to ours independently proposed to
use Batch Normalization statistics to reconstruct input data~\cite{haroush2019knowledge}.
They propose a knowledge-distillation based method to boost the accuracy further, by
generating input data that is similar to the original training dataset, using the so-called Inceptionism~\cite{mordvintsev2015inceptionism}.
However, it is not clear how the latter approach can be used for tasks such as object detection or image segmentation. Furthermore, this knowledge-distillation process adds to the computational time required for zero-shot quantization.
As we will show in our work, it is possible to use batch norm statistics combined with mixed-precision quantization to achieve state-of-the-art accuracy, and importantly
this approach is not limited to image classification task.
In particular, we will present results on object detection using RetinaNet-ResNet50,
besides testing \OURS on 
a wide range of models for image classification (using ResNet18/50/152, MobileNetV2, ShuffleNet, SqueezeNext, and InceptionV3), We show that for all of these cases \OURS exceeds state-of-the-art quantization
performance. 
Importantly, our approach has a very small computational overhead. For example, we can finish ResNet50 quantization in under 30 seconds on an 8 V-100 system (corresponding
to 0.5\% of one epoch training time of ResNet50 on ImageNet).

Directly quantizing all NN layers to low precision can lead to significant accuracy degradation.
A promising approach to address this is to perform mixed-precision 
quantization~\cite{dong2019hawq, dong2019hawqv2,wu2018mixed,zhou2018adaptive,wang2018haq}, where different
bit-precision is used for different layers.
The key idea behind mixed-precision quantization is that not all layers of a convolutional network are equally ``sensitive'' to quantization.
A na\"{\i}ve mixed-precision quantization method can be computationally expensive, as the
search space for determining the precision of each layer is exponential in the number of layers.
To address this, \cite{wang2018haq} uses NAS/RL-based search algorithm to explore the configuration space. 
However, these searching methods can be expensive and are often sensitive to the hyper-parameters and the initialization of the RL based algorithm.
Alternatively, the recent work of~\cite{dong2019hawq,shen2019q,dong2019hawqv2} introduces a Hessian based method, where
the bit precision setting is based on the second-order sensitivity of each layer. 
However, this approach does require access to the original training set, a limitation which we address in \OURS.

%% file: _s3_methods.tex
\section{Methodology}\label{sec:method}
For a typical supervised computer vision task, we seek to minimize the empirical risk loss, i.e., 
\small 
\begin{equation}\label{eq:basic_erl}   
\min_\theta \L(\theta) = \frac1N \sum_{i=1}^N f(\M(\theta; x_i),  y_i),
\end{equation}
\normalsize
where $\theta\in \R^n$ is the learnable parameter, $f(\cdot, \cdot)$ is the loss function (typically cross-entropy loss), $(x_i, y_i)$ is the training input/label pair, $\M$ is the NN model with $L$ layers, and $N$ is the total number of training data points.
Here, we assume that the input data goes          through standard preprocessing normalization of zero mean ($\mu_0=0$) and unit variance ($\sigma_0=1$).
Moreover, we assume that the model has $L$ BN layers
denoted as $BN_1$, $BN_2$, ..., $BN_L$. We denote the activations before the i-th BN layer with  $z_i$ (in other words $z_i$ is the output of the i-th convolutional layer).
During inference,
$z_i$ is normalized by the running mean ($\mu_i$) and variance ($\sigma_i^2$) of parameters in the i-th BN layer ($BN_i$), which is pre-computed during the training process. 
Typically BN layers also include scaling and bias correction, which we denote as $\gamma_i$ and $\beta_i$, respectively.

We assume that before quantization, all the NN parameters and activations are stored in 32-bit precision and
that we have no access to the training/validation datasets. 
To quantize a tensor (either weights or activations), we clip the parameters to a range of
 $[a, b]$ ($a,b \in \R$), and we uniformly discretize the space to $2^k-1$ even intervals using asymmetric 
 quantization. 
That is, the length of each interval will be $\Delta=\frac{b-a}{2^k-1}$.
As a result, the original 32-bit single-precision values are mapped to unsigned integers within the range of $[0, 2^k-1]$.
Some work has proposed non-uniform quantization schemes which can capture finer details
of weight/activation distribution~\cite{park2018value, han2015deep, Zhang_2018_ECCV}.
However, we only use asymmetric uniform quantization, as the non-uniform methods
are typically not suitable for efficient hardware execution.

The \OURS framework supports both fixed-precision and mixed-precision quantization.
In the latter scheme, different  layers of the model could have different bit precisions  (different $k$).
The main idea behind mixed-precision quantization is to keep more sensitive layers at higher
precision, and  more aggressively quantize less sensitive layers, without increasing overall model size.
As we will show later, this mixed-precision quantization is key to achieving high accuracy for
ultra-low precision settings such as 4-bit quantization.
Typical choices for $k$ for each layer are $\{2,4,8\}$ bit. Note that
this mixed-precision quantization leads to exponentially large search space, as every layer could have one 
of these bit precision settings.
It is possible to avoid this prohibitive search space if we could measure the sensitivity of 
the model to the quantization of each layer~\cite{dong2019hawq,shen2019q,dong2019hawqv2}. For the case of post-training quantization (i.e. without fine-tuning),
a good sensitivity metric is to use
Kullback–Leibler (KL) divergence between the original model and the quantized
model, defined~as:

\small 
\begin{equation}\label{eq:sensivity_formula}
    \Omega_i(k) = \frac{1}{N} \sum_{j=1}^{N_{dist}} \textsc{KL}(\M(\theta; x_j), \M(\tilde\theta_i(\textit{k-bit}); x_j)).
\end{equation}
\normalsize
where $\Omega_i(k)$ measures how sensitive the $i$-th layer is when quantized to \textit{k-bit}, and $\tilde\theta_i(\textit{k-bit})$ refers to quantized model parameters in the $i$-th layer
with $k$-bit precision.
If $\Omega_i(k)$ is small, the output of the quantized model will not significantly deviate from the output of the full precision model when quantizing the $i$-th layer to $k$-bits, and thus the $i$-th layer is relatively insensitive to \textit{k-bit} quantization, and vice versa. 
This process is schematically shown in 
\fref{fig:resnet18_sensitivity_illustration} for ResNet18.
However, an important problem is that for zero-shot quantization we do not have access to the
original training dataset $x_j$ in~\eref{eq:sensivity_formula}.
We address this by ``distilling'' a synthetic input data to match the statistics of the original
training dataset, which we refer to as \rg.
We obtain the \rg by solely analyzing the trained model itself, 
as described below.

\subsection{\rg}
\label{sec:synthetic_data}
For zero-shot quantization, we do not have access to any of the training/validation data.
This poses two challenges. 
First, we need to know the range of values for
activations of each layer so that we can clip the range for quantization (the
$[a, b]$ range mentioned above). However, we cannot determine this
range without access to the training dataset. This is a problem for both uniform and mixed-precision quantization.
Second, another challenge is that for mixed-precision quantization, we need to compute  $\Omega_i$ in~\eref{eq:sensivity_formula}, but we do not have access to training data $x_j$.
A very na\"{\i}ve method to address these challenges is to create a random input data
drawn from a Gaussian distribution with zero mean and unit variance and feed it into the model.
However, this approach cannot capture the correct statistics of the activation data corresponding
to the original training dataset. This is illustrated in~\fref{fig:sensitivity_resnet50_mobilenetv2} (left), 
where we plot the sensitivity of each layer of ResNet50 on ImageNet measured with the original training
dataset (shown in black) and Gaussian based input data (shown in red). As one can see, the Gaussian 
data clearly does not capture the correct sensitivity of the model. 
For instance, for the first three layers, the sensitivity order of the red line is actually the opposite
of the original training data.

\begin{figure}[!ht]
\centering
\includegraphics[width=.49\linewidth, trim={10cm 10cm 10cm 10cm}, clip]{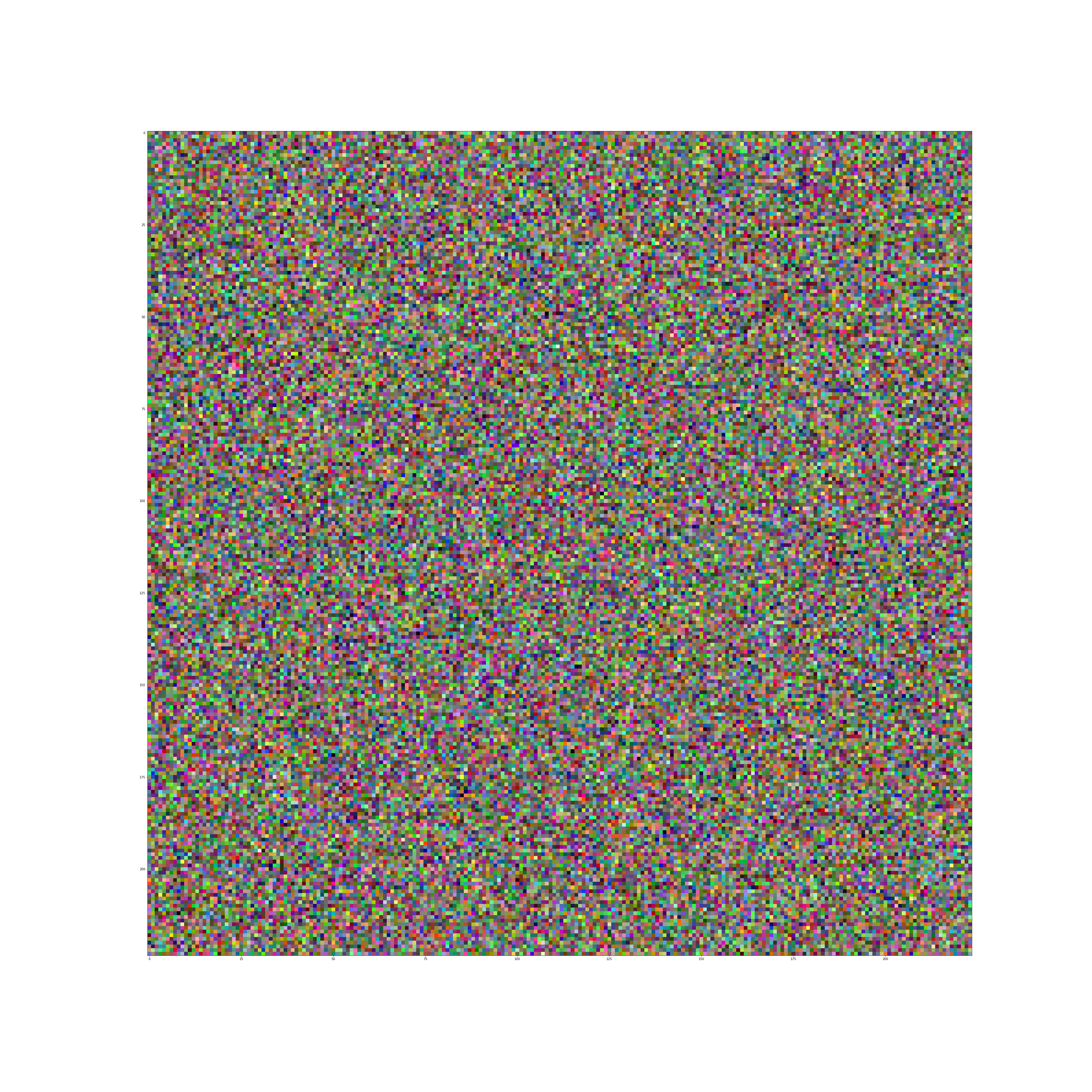}
\includegraphics[width=.49\linewidth, trim={10cm 10cm 10cm 10cm}, clip]{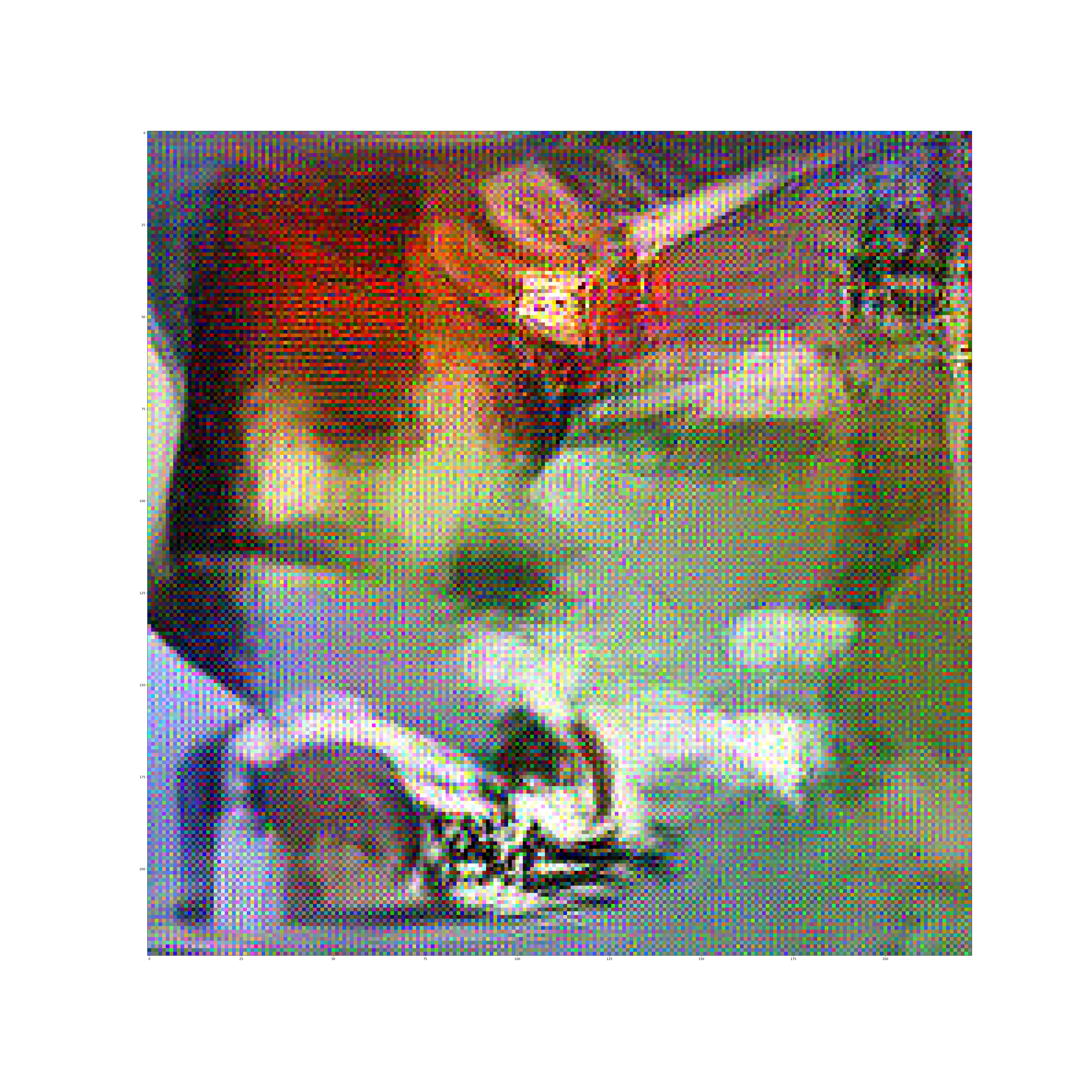}
\caption{Visualization of Gaussian data (left) and \rg (right). More local structure can be seen in our \rg that is generated according to~\aref{alg:distilled_data}. }
\label{fig:visualization_synd}
\end{figure}

\normalsize
To address this problem, we propose a novel method to ``distill'' input data from the NN model itself, i.e., to generate synthetic data carefully engineered based on the properties of the NN.
In particular, we solve a distillation optimization problem, in order to learn an input data distribution
that best matches the statistics encoded in the BN layer of the model. 
In more detail, we solve the following
optimization problem:

\small 
\begin{equation}\label{eq:synthetic_gaussian}
    \min_{x^r} \sum_{i=0}^L \|\tilde \mu_i^r - \mu_i\|_2^2 + \|\tilde \sigma_i^r - \sigma_i\|_2^2,
\end{equation}
\normalsize
where $x^r$ is the reconstructed (distilled) input data, and $\mu_i^r$/$\sigma_i^r$ are the mean/standard deviation of the 
\rg's distribution at layer $i$, and $\mu_i$/$\sigma_i$ are the corresponding mean/standard deviation parameters stored in the BN layer at layer $i$. 
In other words, after solving this optimization problem, we can distill an input data which, when fed into the network, can have a statistical distribution that closely matches the original model.
Please see~\aref{alg:distilled_data} for a description. 
This \rg can then be used to address the two challenges described earlier.
First, we can use the \rg's activation range to determine quantization clipping parameters (the $[a, b]$ range mentioned above).
Note that some prior work~\cite{banner2018post,lee2018quantization,zhao2019improving} address this by using limited (unlabeled) data to determine the activation range.
However, this contradicts the assumptions of zero-shot quantization, and
may not be applicable for certain applications.
Second, we can use the \rg and feed it in~\eref{eq:sensivity_formula} to determine the quantization sensitivity ($\Omega_i$).
The latter is plotted for ResNet50 in~\fref{fig:sensitivity_resnet50_mobilenetv2} (left) shown in solid blue 
color. As one can see, the \rg closely matches the sensitivity of the model as compared
to using Gaussian input data (shown in red). 
We show a visualization of the random Gaussian data as well as the \rg for ResNet50 in~\fref{fig:visualization_synd}. We can see that the \rg can capture fine-grained local structures.

\input{_distill_data_alg.tex}

\subsection{Pareto Frontier}
\label{sec:pareto_frontier}

\begin{figure}[!ht]
\centering
\includegraphics[width=\linewidth]{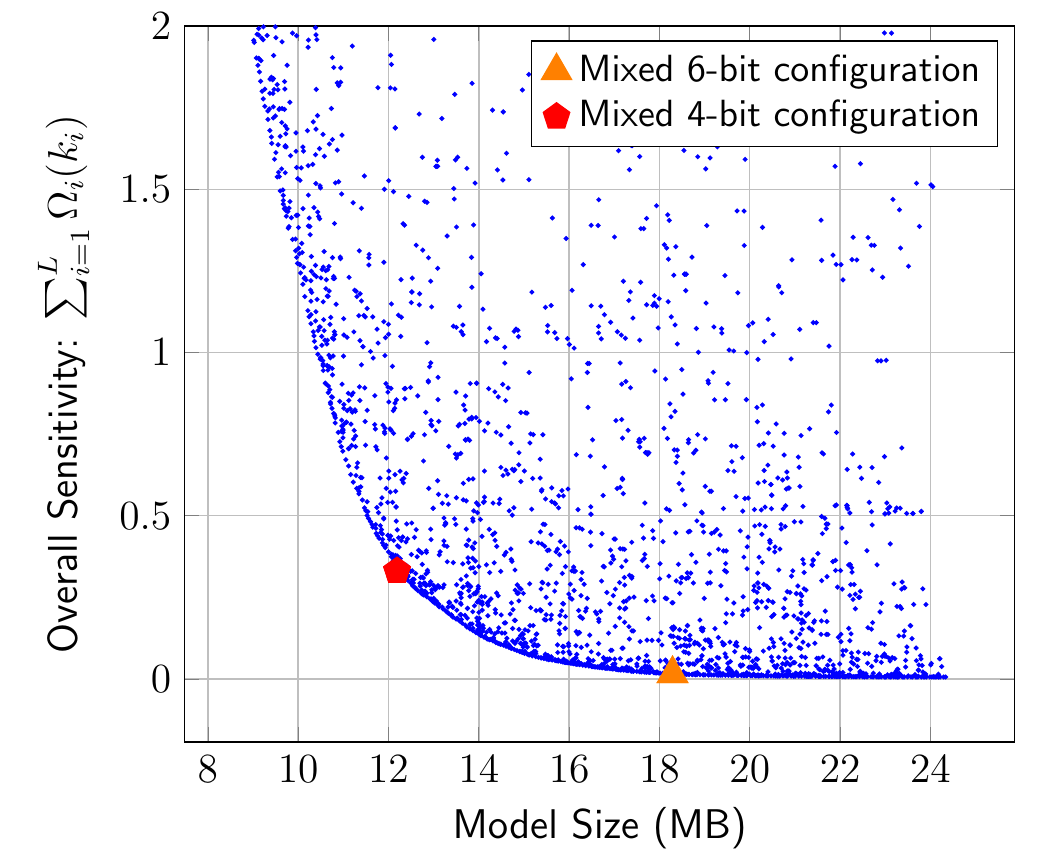}
\caption{
The Pareto frontier of ResNet50 on ImageNet. 
Each point shows a mixed-precision bit setting. The x-axis
shows the resulting model size for each configuration, and the y-axis
shows the resulting sensitivity.
In practice, a constraint for model size is set.
Then the Pareto frontier method chooses a bit-precision configuration
that results in minimal perturbation. We show two examples
for 4 and 6-bit mixed precision configuration shown in red and orange.
The corresponding results are presented in~\tref{tab:resnet50}. 
}
  \label{fig:pareto_frontier}
\end{figure}

As mentioned before, the main challenge for mixed-precision quantization is to determine the exact bit precision configuration for the entire NN. 
For an L-layer model with $m$ possible precision options, the mixed-precision search space, denoted as $\mathcal{S}$, has an exponential size of $m^L$. 
For example for ResNet50 with just three bit precision of  $\{2,4,8\}$ (i.e., $m=3$), the search space contains $7.2\times 10^{23}$ configurations.
However, we can use the sensitivity metric in~\eref{eq:sensivity_formula} to reduce this search space.
The main idea is to use higher bit precision for layers that are more sensitive, and lower bit precision for layers that are less sensitive. 
This gives us a relative ordering on the number of bits. 
To compute the precise bit precision setting, we propose a Pareto frontier approach similar to the method used in~\cite{dong2019hawqv2}.

The Pareto frontier method works as follows. For a target quantized model size of  $S_{target}$, we measure the overall sensitivity of the model for each bit precision configuration that results in the $S_{target}$ model size. 
We choose the bit-precision setting that corresponds to the minimum overall sensitivity. 
In more detail, we solve the following optimization problem:
\small 
\begin{equation}\label{eq:sensitivity_objective}
    \min_{\{k_i\}_{i=1}^L} \Omega_{sum} = \sum_{i=1}^L \Omega_i(k_i)~~s.t.~\sum_{i=1}^L P_i*k_i \leq S_{target},
\end{equation}
\normalsize
where $k_i$ is the quantization precision of the i-th layer, and $P_i$ is the parameter size for the $i$-th layer. Note that here we make the
simplifying assumption
that the sensitivity of different layers are independent of 
the choice of bits for other layers (hence $\Omega_i$ only depends
on the bit precision for the $i$-th layer).\footnote{Please see~\sref{sec:advanced_pareto_frontier} where we describe how
we relax this assumption without having to perform
an exponentially large computation for the sensitivity for each bit precision
setting.}
Using a dynamic programming method we can solve the best setting with different $S_{target}$ together, and then we plot the Pareto frontier. 
An example is shown in~\fref{fig:pareto_frontier} for ResNet50 model, where the x-axis
is the model size for each bit precision configuration, and the y-axis is the overall model perturbation/sensitivity.
Each blue dot in the figure represents a mixed-precision configuration.
In \OURS, we choose the bit precision setting that has the smallest
perturbation with a specific model size constraint.

Importantly, note that the computational overhead of computing the Pareto frontier is $\mathcal{O}(mL)$. 
This is because we compute the sensitivity of each layer separately from other layers. 
That is, we compute sensitivity $\Omega_i$ ($i=1,2,...,L$) with respect to all $m$ different precision options, which leads to the  $\mathcal{O}(mL)$ computational complexity. 
We should note that this Pareto Frontier approach (including the Dynamic Programming optimizer), is not theoretically guaranteed to result in the best possible configuration, out of all possibilities in the exponentially large search space.
However, our results show that the final mixed-precision configuration achieves state-of-the-art accuracy with small performance loss, as compared to the original model in single precision.
 

%% file: _distill_data_alg.tex
\begin{algorithm}[t]
\DontPrintSemicolon
\caption{Generation of Distilled Data}
\label{alg:distilled_data}
    \SetAlgoLined
    \KwInput{ 
    Model: $\M$ with $L$ Batch Normalization layers
    }
    
    \KwOutput{
    A batch of distilled data: $x^r$ 
    }
    
     Generate random data from Gaussian: $x^r$ 
    
    Get $\mu_i, \sigma_i$ from Batch Normalization layers of $\M$, $i\in{0,~1,~\dots,~L}$ \tcp*{Note that $\mu_0=0,~\sigma_0=1$} 
    
    \For{j $=1,2,\ldots$}{
        Forward propagate $\M(x^r)$ and gather intermediate activations
        
        Get $\tilde\mu_i$ and $\tilde\sigma_i$ from intermediate activations, $i\in{1,~\dots,~n}$ 
        
        Compute $\tilde\mu_0$ and $\tilde\sigma_0$ of $x^r$
        
        Compute the loss based on~\eref{eq:synthetic_gaussian} 
        
        Backward propagate and update $x^r$
    }
    
    
\end{algorithm}

%% file: _s4_results.tex
\begin{table}[!ht]
\caption{Quantization results of ResNet50, MobileNetV2, and ShuffleNet on ImageNet. 
We abbreviate quantization bits used for weights as ``W-bit'' (for activations as ``A-bit''),
top-1 test accuracy as ``Top-1.''
Here, ``MP'' refers to mixed-precision quantization,
``No D'' means that none of the data is used to assist quantization, 
and ``No FT'' stands for no fine-tuning (re-training). 
Compared to post-quantization methods OCS~\cite{zhao2019improving}, OMSE~\cite{Kravchik_2019_ICCV}, and DFQ~\cite{nagel2019data}, \OURS achieves better accuracy.
$\text{\OURS}^\dagger$ means using percentile for quantization.
}
\label{tab:imagenet_main_text}

\subfloat[\footnotesize ResNet50]{
\centering
\centering
\small
\setlength\tabcolsep{1.3pt}
\begin{tabular}{p{8em} ccccccccccccccccccccccccccccc} \toprule
    Method      &{No D}&{No FT}                 &W-bit&A-bit&Size (MB) &Top-1\\
    \midrule
\ha Baseline                        &--&--      &32&32      &97.49              &77.72          \\
\midrule    
\ha OMSE~\cite{Kravchik_2019_ICCV}    &\xm&\xm    &4 &32      &12.28              &70.06          \\
\ha OMSE~\cite{Kravchik_2019_ICCV}    &\cm&\xm    &4 &32      &12.28              &74.98          \\
\ha PACT~\cite{choi2018pact}        &\cm&\cm    &4 &4       &12.19              &76.50          \\
\hc \OURS                           &\xm&\xm    &\AV &8     &\textbf{12.17}     &\textbf{75.80} \\
\hc $\text{\OURS}^\dagger$          &\xm&\xm    &\AV &8     &\textbf{12.17}     &\textbf{76.08} \\
\midrule    
\ha OCS~\cite{zhao2019improving}    &\cm&\xm    &6 &6       &18.46              &74.80          \\
\hc \OURS                           &\xm&\xm    &\AV &6     &\textbf{18.27}     &\textbf{77.43} \\
\midrule    
\hc \OURS                           &\xm&\xm    &8 &8       &{24.37}            &\textbf{77.67} \\
\bottomrule 
\end{tabular}
\label{tab:resnet50}
}

\subfloat[\footnotesize MobileNetV2]{
\centering
\small
\setlength\tabcolsep{1.3pt}
\begin{tabular}{p{8em} ccccccccccccccccccccccccccccc} \toprule
    Method      &{No D}&{No FT} &W-bit&A-bit        &Size (MB) &Top-1\\
    \midrule
\ha Baseline                        &--&--          &32&32      &13.37          &73.03          \\
\midrule
\hc \OURS                           &\xm&\xm        &\AV &8     &{1.67}         &\textbf{68.83} \\
\hc $\text{\OURS}^\dagger$          &\xm&\xm        &\AV &8     &{1.67}         &\textbf{69.44} \\
\midrule
\ha Integer-Only~\cite{jacob2018quantization}&\cm&\cm  &6 &6    &2.50           &70.90          \\
\hc \OURS                           &\xm&\xm        &\AV &6     &{2.50}         &\textbf{72.85} \\
\midrule
\ha RVQuant~\cite{park2018value}    &\cm&\cm        &8&8        &3.34           &70.29          \\
\ha DFQ~\cite{nagel2019data}        &\xm&\xm        &8&8        &3.34           &71.20          \\
\hc \OURS                           &\xm&\xm        &8&8        &{3.34}         &\textbf{72.91} \\
\bottomrule 
\end{tabular}
\label{tab:mobilenetv2}
}

\subfloat[\footnotesize ShuffleNet]{
\centering
\centering
\small
\setlength\tabcolsep{1.3pt}
\begin{tabular}{p{8em}cccccccc} \toprule
    Method      &{No D}&{No FT} &W-bit&A-bit    &Size (MB) &Top-1\\
    \midrule
\ha Baseline    &--&--              &32&32      &5.94               &65.07              \\
\midrule
\hc \OURS       &\xm&\xm            &\AV &8     &{0.74}             &\textbf{58.96}     \\
\midrule    
\hc \OURS       &\xm&\xm            &\AV &6     &{1.11}             &\textbf{62.90}     \\
\midrule    
\hc \OURS       &\xm&\xm            &8 &8       &{1.49}             &\textbf{64.94}     \\
\bottomrule 
\end{tabular}
\label{tab:shufflenet}
}
\end{table}

\section{Results}\label{sec:results}
In this section, we extensively test \OURS on a wide range of models and datasets.
We first start by discussing the zero-shot quantization of ResNet18/50, MobileNet-V2, and ShuffleNet on ImageNet in~\sref{sec:imagenet_result}. 
Additional results for quantizing ResNet152, InceptionV3, and SqueezeNext on ImageNet, as well as ResNet20 on Cifar10 are provided in Appendix~\ref{sec:extra_imagenet}. 
We also present results for object detection using RetinaNet tested on Microsoft COCO dataset in~\sref{sec:objective_detection_result}. 
We emphasize that all of the results achieved by \OURS are 100\% zero-shot without any need for fine-tuning.

We also emphasize that we used exactly the same
hyper-parameters (e.g., the number of iterations to generate \rg) for all experiments, including the results on Microsoft COCO dataset.

\subsection{ImageNet}\label{sec:imagenet_result}
We start by discussing the results on the ImageNet dataset.
For each model, after generating \rg based on~\eref{eq:synthetic_gaussian}, we compute the sensitivity of each layer using~\eref{eq:sensivity_formula} for different bit precision.
Next, we use~\eref{eq:sensitivity_objective} and the Pareto frontier introduced in~\sref{sec:pareto_frontier} to get the best bit-precision configuration based on the overall sensitivity for a given model size constraint. 
We denote the quantized results as WwAh where w and h denote the bit precision used for weights and
activations of the NN~model.

We present zero-shot quantization results for ResNet50 in~\tref{tab:resnet50}.
As one can see, for W8A8 (i.e., 8-bit quantization for both weights and
activations), \OURS results in only 0.05\% accuracy degradation.
Further quantizing the model to W6A6, \OURS achieves 77.43\% accuracy, which is 
2.63\% higher than OCS~\cite{zhao2019improving}, even though our model is slightly smaller (18.27MB as compared to 18.46MB for OCS).%
\footnote{Importantly note that OCS requires access to the training data, while \OURS does not use any training/validation data.}
We show that we can further quantize ResNet50 down to just 12.17MB with mixed precision quantization, 
and we obtain 75.80\% accuracy. Note that this is 0.82\% higher than OMSE~\cite{Kravchik_2019_ICCV} with access to training data and 5.74\% higher than zero-shot version of OMSE. Importantly, note that OMSE keeps
activation bits at 32-bits, while for this comparison our results use 8-bits for the activation (i.e., $4\times$ smaller activation memory footprint than OMSE).
For comparison, we include results for PACT~\cite{choi2018pact}, a standard quantization method that requires access to training data and also requires fine-tuning. 

An important feature of the \OURS framework is that it can perform the quantization with very low
computational overhead. For example, the end-to-end quantization of ResNet50 takes less than 30 seconds
 on an 8 Tesla V100 GPUs (one epoch training time on this system takes 100 minutes).
In terms of timing breakdown, it takes 3s to generate the \rg, 12s to compute the sensitivity for all layers of ResNet50, and 14s to perform Pareto Frontier optimization.

We also show \OURS results on MobileNetV2 and compare it with both DFQ~\cite{nagel2019data} and fine-tuning based methods~\cite{park2018value,jacob2018quantization}, as shown in~\tref{tab:mobilenetv2}. 
For W8A8, \OURS has less than 0.12\% accuracy drop as compared to baseline, and it achieves 1.71\% higher
accuracy as compared to DFQ method. 

Further compressing the model to W6A6 with mixed-precision quantization for weights, \OURS can still outperform Integer-Only~\cite{jacob2018quantization} by 1.95\% accuracy, even though \OURS does not use any data or fine-tuning. 
\OURS can achieve 68.83\% accuracy even when the weight compression is 8$\times$, which corresponds to using 4-bit quantization for weights on~average.

We also experimented with percentile based clipping to determine the quantization range~\cite{li2019fully} (please see~\sref{sec:clipping} for details). The results corresponding to percentile based clipping
are denoted as $ZeroQ^\dagger$ and reported in~\tref{tab:imagenet_main_text}. We found
that using percentile based clipping is helpful for low precision quantization.
Other choices for clipping methods have been proposed in the literature.
Here we note that our approach is orthogonal to these improvements and that \OURS could
be combined with these methods.

We also apply \OURS to quantize efficient and highly compact models such as ShuffleNet, whose model size is only 5.94MB.
To the best of our knowledge, there exists no prior zero-shot quantization results for this model.
\OURS  achieves a small accuracy drop of 0.13\% for W8A8.
We can further quantize the model down to an average of 4-bits for weights, which achieves a model size of
only 0.73MB, with an accuracy of 58.96\%.

\begin{table}[!htbp]
\caption{Object detection on Microsoft COCO using RetinaNet. 
By keeping activations to be 8-bit, our 4-bit weight result is comparable with recently proposed method~FQN~\cite{li2019fully}, which relies on fine-tuning.
(Note that FQN uses 4-bit activations and the baseline used in~\cite{li2019fully} is 35.6~mAP).} 
\label{tab:Detection}
\centering
\small
\setlength\tabcolsep{2 pt}
\begin{tabular}{p{6em}ccccccccccccccccccccccccccccc} \toprule
    Method      &No D&No FT    &W-bit&A-bit    &Size (MB)          &mAP\\
    \midrule
\ha Baseline                &\xm&\xm        &32&32          &145.10             &36.4          \\
\midrule
\ha FQN~\cite{li2019fully}  &\cm&\cm        &4&4  &18.13              &32.5\\
\hc \OURS                   &\xm&\xm        &\AV&8          &{18.13}            &\textbf{33.7} \\
\midrule
\hc \OURS                   &\xm&\xm        &\AV&6          &{24.17}            &\textbf{35.9} \\
\midrule
\hc \OURS                   &\xm&\xm        &8&8            &{36.25}            &\textbf{36.4} \\
\bottomrule 
\end{tabular}
\end{table}

We also compare with the recent Data-Free Compression (DFC)~\cite{haroush2019knowledge} method.
There are two main differences between \OURS and DFC. 
First, DFC proposes a fine-tuning method to recover accuracy for ultra-low precision cases. 
This can be time-consuming and as we show it is not necessary. 
In particular, we show that with mixed-precision quantization one can actually achieve higher accuracy without any need for fine-tuning. 
This is shown in~\tref{tab:resnet18_dfc} for ResNet18 quantization on ImageNet.
In particular, note the results for W4A4, where the DFC method without fine-tuning results in more than 15\% accuracy drop with a final accuracy of 55.49\%. 
For this reason, the authors propose a method with post quantization training, which can boost the accuracy to 68.05\% using W4A4 for intermediate layers, and 8-bits for the first and last layers.
In contrast, \OURS achieves a higher accuracy of 69.05\% without any need for fine-tuning.
Furthermore, the end-to-end zero-shot quantization of ResNet18 takes only 12s on an 8-V100 system (equivalent to $0.4\%$ of the 45 minutes time for one epoch training of ResNet18 on ImageNet).
Secondly, DFC method uses Inceptionism~\cite{mordvintsev2015inceptionism} to facilitate the generation of data with random labels, but it is hard to extend this for object detection and image segmentation tasks.

\begin{table}[!htbp]
\caption{Uniform post-quantization on ImageNet with ResNet18. We use percentile clipping for W4A4 and W4A8 settings. $\text{\OURS}^\dagger$ means using percentile for quantization.}
\label{tab:resnet18_dfc}
\centering
\small
\setlength\tabcolsep{2 pt}
\begin{tabular}{p{8em}ccccccccccccccccccccccccccccc} \toprule
    Method                          &{No D}&{No FT} &W-bit&A-bit    &Size (MB)      &Top-1\\
    \midrule
\ha Baseline                        &--&--          &32&32          &44.59          &71.47      \\
\midrule
\ha PACT~\cite{choi2018pact}        &\cm&\cm        &4 &4           &5.57           &69.20      \\
\ha DFC~\cite{haroush2019knowledge} &\xm&\xm        &4 &4           &5.58           &55.49\\
\ha DFC~\cite{haroush2019knowledge} &\xm&\cm        &4 &4           &5.58           &68.06\\
\hc \OURS                           &\xm&\xm        &\AV&4          &\textbf{5.57}  &\textbf{-- }\\
\hc $\text{\OURS}^\dagger$          &\xm&\xm        &\AV&4          &\textbf{5.57}  &\textbf{69.05}\\
\midrule
\ha Integer-Only\cite{jacob2018quantization}&\cm&\cm&6 &6           &8.36           &67.30      \\
\ha DFQ~\cite{nagel2019data}        &\xm&\xm        &6 &6           &8.36           &66.30      \\
\hc \OURS                           &\xm&\xm        &\AV&6          &\textbf{8.35}  &\textbf{71.30}\\
\midrule
\ha RVQuant~\cite{park2018value}    &\cm&\cm        &8&8            &11.15          &70.01      \\
\ha DFQ~\cite{nagel2019data}        &\xm&\xm        &8&8            &11.15          &69.70      \\
\ha DFC~\cite{haroush2019knowledge} &\xm&\cm        &8&8            &11.15          &69.57\\
\hc \OURS                           &\xm&\xm        &8&8            &{11.15}        &\textbf{71.43}\\
\bottomrule 
\end{tabular}
\end{table}

We include additional results of quantized ResNet152, InceptionV3, and SqueezeNext on ImageNet, as well as ResNet20 on Cifar10, in Appendix~\ref{sec:extra_imagenet}.

\subsection{Microsoft COCO}
\label{sec:objective_detection_result}
Object detection is often much more complicated than ImageNet classification. To demonstrate the flexibility of our approach we also test \OURS on an object detection task on Microsoft COCO dataset. 
RetinaNet~\cite{lin2017focal} is a state-of-the-art single-stage detector,
and we use the pretrained model with ResNet50 as the backbone, which can achieve 36.4~mAP.%
\footnote{Here we use the standard mAP 0.5:0.05:0.95 metric on COCO dataset.}

One of the main difference of RetinaNet with previous NNs we tested on ImageNet is that some convolutional layers in RetinaNet are not followed by BN layers.
This is because of the presence of a feature pyramid network (FPN)~\cite{DBLP:journals/corr/LinDGHHB16}, and it means that the number of BN layers is slightly smaller than that of convolutional layers. 
However, this is not a limitation and the \OURS framework still works well.
Specifically, we extract the backbone of RetinaNet and create \rg.
Afterwards, we feed the \rg into RetinaNet to measure the sensitivity as well as to determine the activation range for the entire NN. 
This is followed by optimizing for the Pareto Frontier, discussed earlier. 

The results are presented in~\tref{tab:Detection}.
We can see that for W8A8 \OURS has no performance degradation. 
For W6A6, \OURS achieves 35.9 mAP.
Further quantizing the model to an average of 4-bits for the weights, \OURS achieves 33.7 mAP.
Our results are comparable to the recent results of FQN~\cite{li2019fully}, even
though it is not a zero-shot quantization method (i.e., it uses the full training dataset and requires fine-tuning). 
However, it should be mentioned that \OURS keeps the activations to be 8-bits, while FQN uses 4-bit activations.

%% file: _s5_ablation.tex
\section{Ablation Study}\label{sec:ablation_study}
Here, we present an ablation study for the two components of \OURS:
(i) the \rg generated by~\eref{eq:synthetic_gaussian} to help sensitivity analysis and determine activation clipping range; and
(ii) the Pareto frontier method for automatic bit-precision assignment.
Below we discuss the ablation study for each part separately.

\subsection{\rg}
In this work, all the sensitivity analysis and the activation range are computed on the \rg.
Here, we perform an ablation study on the effectiveness of \rg as compared to using just Gaussian data.
We use three different types of data sources, (i) Gaussian data with mean ``0'' and variance ``1'', (ii) data from training dataset, (iii) our \rg, as the input data to measure the sensitivity and to determine the activation range. 
We quantize ResNet50 and MobileNetV2 to an average of 4-bit for weights and 8-bit for activations, and we report results in~\tref{tab:ablation_recon}. 

For ResNet50, using training data results in 75.95\% testing accuracy. 
With Gaussian data, the performance degrades to 75.44\%. 
\OURS can alleviate the gap between Gaussian data and training data and achieves 75.80\%. 
For more compact/efficient models such as MobileNetV2, the gap between using Gaussian data and using training data increases to 2.33\%. 
\OURS can still achieve 68.83\%, which is only 0.23\% lower than using training data. 
Additional results for ResNet18, ShuffleNet and SqueezeNext are shown in~\tref{tab:Abalation_SynD_extra}.

\begin{table}[!htbp]
\caption{Ablation study for \rg on ResNet50 and MobileNetv2.  
We show the performance of \OURS with different data to compute the sensitivity and to determine the activation range. 
All quantized models have the same size as models with 4-bit weights and 8-bit activations.
}
\label{tab:ablation_recon}
\centering
\small
\setlength\tabcolsep{2.5pt}
\begin{tabular}{lcccccccccccccccccccccccccccccccccccccccccc} \toprule
Method              & W-bit & A-bit &ResNet50   &MobileNetV2\\
\midrule
\ha Baseline        &32     &32        &77.72  &73.03\\ 
\midrule
\ha Gaussian        &\AV&8          &75.44 &66.73\\
\ha Training Data   &\AV&8          &75.95 &69.06\\
\midrule
\hc \rg           &\AV&8          &\textbf{75.80} &\textbf{68.83}\\
\bottomrule 
\end{tabular}
\end{table}

\subsection{Sensitivity Analysis}
Here, we perform an ablation study to show that the bit precision of
the Pareto frontier method works well. To test this, we compare \OURS with two cases, one where we
choose a bit-configuration that corresponds to maximizing $\Omega_{sum}$ (which is opposite
to the minimization that we do in \OURS), and one case where we use random bit precision for different layers.
We denote these two methods as Inverse and Random.
The results for quantizing weights to an average of 4-bit and activations to 8-bit are shown in~\tref{tab:Abalation_Pareto}. 
We report the best and worst testing accuracy as well as the mean and variance in the results out of 20 tests.
It can be seen that \OURS results in significantly better testing performance as compared to Inverse and Random.
Another noticeable point is that the best configuration (i.e., minimum $\Omega_{sum}$) can outperform 0.18\% than the worst case among the top-20 configurations from \OURS, which reflects the advantage of the Pareto frontier method. 
Also, notice the small variance of all configurations generated by \OURS.

\begin{table}[!htbp]
\caption{Ablation study for sensitivity analysis on ImageNet (W4A8) with ResNet50. Top-20 configurations are selected based on different sensitivity metric types. We report the best, mean, and worst accuracy among 20 configurations. ``\OURS'' and ``Inverse'' mean selecting the bit configurations to minimize and maximize the overall sensitivity, respectively, under the average 4-bit weight constraint. ``Random'' means randomly selecting the bit for each layer and making the total size equivalent to 4-bit weight quantization.}
\label{tab:Abalation_Pareto}
\centering
\small
\setlength\tabcolsep{5.0pt}
\begin{tabular}{l|cccccccccccccccccccccccccccccccccccccccccc} \toprule
                    &\multicolumn{4}{c}{Top-1 Accuracy}\\
\midrule
\ha Baseline        &\multicolumn{4}{c}{77.72}\\
\ha Uniform         &\multicolumn{4}{c}{66.59}\\
\midrule
   &Best   &Worst  &Mean   &Var\\
\midrule  
\ha Random          &38.98  &0.10   &6.86   &105.8\\ 
\ha Inverse    &0.11   &0.06   &0.07   &3.0$\times10^{-4}$\\
\hc \OURS           &\textbf{75.80}  &\textbf{75.62}  &\bf{75.73}  &\bf{2.4}$\bf{\times10^{-3}}$\\
\bottomrule 
\end{tabular}
\end{table}

%% file: _s6_conclusions.tex
\section{Conclusions}
\label{sec:conclusion}

We have introduced \OURS, a novel post-training quantization method that does not require any access to the training/validation data.
Our approach uses a novel method to distill an input data distribution
to match the
statistics in the batch normalization layers of the model.
We show that this \rg is very effective in capturing the sensitivity of different layers
of the network.
Furthermore, we present a Pareto frontier method to 
select automatically the bit-precision configuration for mixed-precision settings.
An important aspect of \OURS is its low computational overhead. 
For example, the end-to-end zero-shot quantization time of ResNet50 is less than 30 seconds on an 8-V100 GPU system.
We extensively test \OURS on various datasets and models.  This includes various ResNets, InceptionV3, MobileNetV2, ShuffleNet, and SqueezeNext on ImageNet, ResNet20 on Cifar10, and even RetinaNet for
object detection on Microsoft COCO dataset.
We consistently achieve higher accuracy with the same or smaller model size compared to previous post-training quantization methods.
All results show that \OURS could exceed previous zero-shot quantization methods. 
We have open sourced \OURS framework~\cite{zeroq}.

%% file: _s7_appendix.tex
\begin{figure*}[!htbp]
\centering
\includegraphics[width=.9\textwidth]{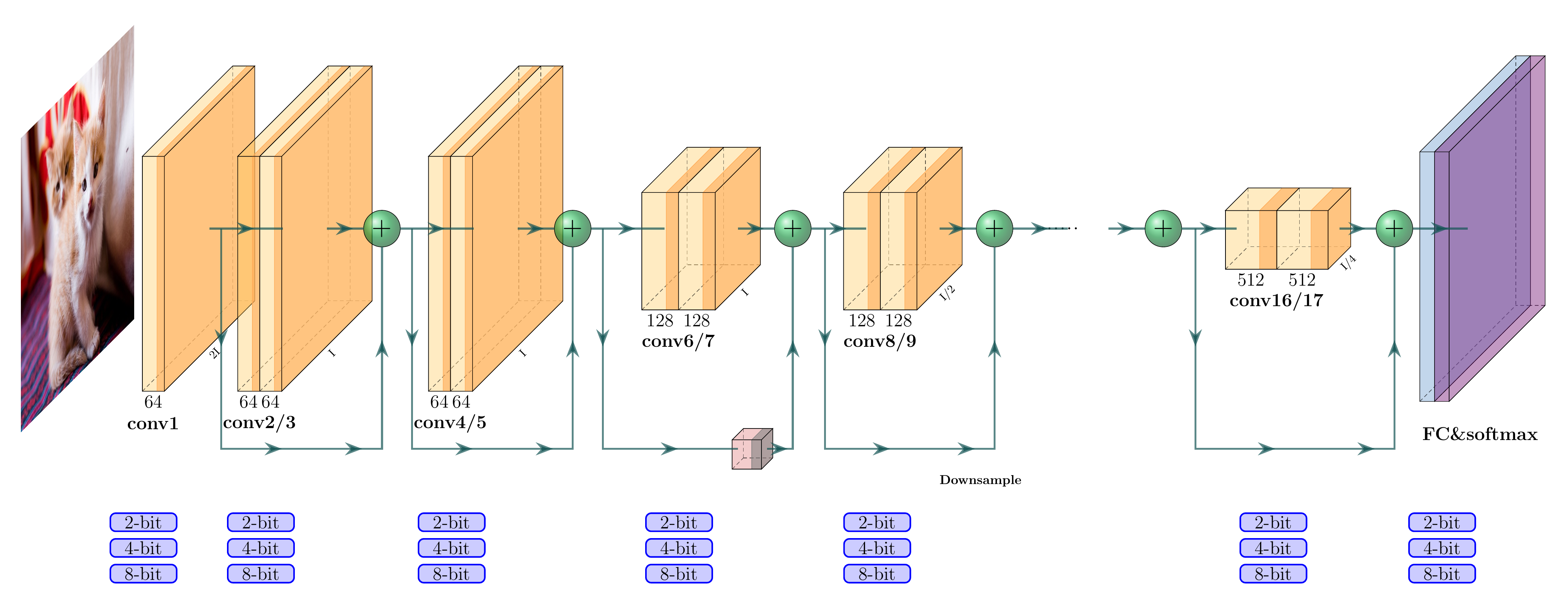}
\caption{
  Mixed precision illustration of ResNet18 on ImageNet.
}
  \label{fig:resnet18_mixedprecision_illustraction}
\end{figure*}

\subsection{Pareto Frontier}
\label{sec:advanced_pareto_frontier}

In~\sref{sec:pareto_frontier}, we presented
how we compute the overall sensitivity incurred by performing mixed-precision
quantization. In particular, in~\eref{eq:sensitivity_objective} we made the simplifying assumption that the sensitivity of each layer to quantization is independent to sensitivity of other layers (we refer to this as independence assumption). This is clearly not the case in practice. One can instead directly compute the sensitivity for 
each possible bit-precision computation without any approximation but this 
is not possible as there are $m^L$ possible bit-precision configurations.
Here we discuss our approach which falls in between these two extremes.
Instead of computing the sensitivity of the entire network at once, we 
break the network into $L/a$ groups, with each group containing $a$ layers.
Furthermore, we break the x-axis (model size) of the Pareto frontier plot into $b$ intervals in every steps mentioned below.

We start with the first $a$ layers of the network. We compute the sensitivity
of these layers with the independence assumption. This means we only have
to compute $m \times a$ sensitivities. Afterwards, for each interval on the x-axis, we 
choose top $\tilde{t}$ configurations that have the lowest overall sensitivity
when the first $a$ layers are quantized. We then relax the independence assumption
for these $\tilde{t}$ configurations and recompute the overall sensitivity, $\Omega_{sum}$, without any approximation. This leads to a cost of $\tilde{t}$
of computing~\eref{eq:sensitivity_objective}. We then select the top $t$ configurations
out of these. We conduct this process for all the $b$ intervals for the model size.
Therefore, the total cost will be $\tilde{t} \times b$.

The next step is to consider the next set of $a$ layers. This is similar to
the algorithm for the first step, except that now we need to consider the
top $t \times b$ configurations selected for the first $a$ layers. We first make the independence
assumption for the second $a$ layers. Then we choose top $\tilde{t} \times b$ configurations out of all $t \times b \times m^a$ possible bit configurations for the first $2a$ layers (this number is obtained by combining the top $t \times b$ configurations of the first $a$ layers
and $m^a$ possible configurations in the second $a$ layers). Similar to before,
we then relax the independence assumption and compute the correct
sensitivity, $\Omega_{sum}$, without any approximation. We then select
the final top $t \times b$ configurations for the first $2a$ layers based on this. 

This process needs to be performed for all the $L/a$ groups. As a result, the
total computational cost becomes $(L/a)\times \tilde{t} \times b + m \times L$.
We find that this approach gives a good trade-off between the two
extremes. Our experiments show that the accuracy is not sensitive to the hyperparameters, and we typically set $\tilde{t}$, $t$, $b$, $a$ to be 10, 5, 200, 5, respectively. It should be noted that this approach has a small computational overhead but can automatically lead to bit precision settings with good empirical results. 

\subsection{Results on CIFAR-10}
In this section, we show the results of our \OURS on CIFAR-10 dataset with ResNet20. 
See Table~\ref{tab:cifar_appendix}.
\begin{table}[!htbp]
\caption{ResNet20 on CIFAR-10}
\label{tab:cifar_appendix}
\centering
\setlength\tabcolsep{2.5pt}
\centering
\small
\begin{tabular}{lccccccccccccccccccccccccccccc} \toprule
    Method      &{No Data}&{No FT} &W-bit&A-bit     &Size (MB)          &Top-1\\
    \midrule
\ha Baseline    &--&--              &32&32          &1.04               &94.03      \\
\midrule    
\hc \OURS       &\xm&\xm            &\AV &8         &{0.13}      &\textbf{93.16}\\
\midrule    
\hc \OURS       &\xm&\xm            &\AV &6         &{0.20}      &\textbf{93.87}\\
\midrule
\hc \OURS       &\xm&\xm            &8 &8           &{0.26}      &\textbf{93.94}\\
\bottomrule 
\end{tabular}
\end{table}

\subsection{Extra Results on ImageNet}
\label{sec:extra_imagenet}
In this section, we show extra results for our \OURS on ImageNet with ResNet152, InceptionV3, and SqueezeNext in~\tref{tab:imagenet_appendix}. 
We also show more results to illustrate the effect of \rg compared with Gaussian noise in~\tref{tab:Abalation_SynD_extra}.
\begin{table*}[!htbp]
\centering
\caption{Additional results on ImageNet}
\label{tab:imagenet_appendix}
\subfloat[\footnotesize ResNet152]{
\centering
\centering
\small
\setlength\tabcolsep{2pt}
\begin{tabular}{p{8em}ccccccccccccccccccccccccccccc} \toprule
    Method      &{No D}&{No FT} &w-bit&a-bit        &Size (MB)          &Top-1          \\
\midrule
\ha Baseline    &--&--              &32&32          &229.62             &80.08          \\
\midrule
\hc \OURS       &\xm&\xm            &\AV &8         &{28.70}     &\textbf{78.00} \\
\midrule    
\hc \OURS       &\xm&\xm            &\AV &6         &{43.05}     &\textbf{77.88} \\
\midrule    
\ha RVQuant~\cite{park2018value} &\cm&\cm &8&8      &57.41              &78.35          \\
\hc \OURS       &\xm&\xm            &8 &8           &{57.41}     &\textbf{78.94} \\
\bottomrule 
\end{tabular}
}
\subfloat[\footnotesize InceptionV3]{
\centering
\centering
\small
\setlength\tabcolsep{2pt}
\begin{tabular}{p{8em}ccccccccccccccccccccccccccccc} \toprule
    Method      &{No D}&{No FT} &W-bit&A-bit        &Size (MB)              &Top-1          \\
    \midrule
\ha Baseline    &--&--              &32&32          &90.92                  &78.88          \\
\midrule    
\hc \OURS       &\xm&\xm            &\AV &8         &\bf{11.35}         &\textbf{77.57} \\
\midrule    
\ha OCS\cite{zhao2019improving} &\cm&\xm&6&6        &17.22                  &71.30          \\
\hc \OURS       &\xm&\xm            &\AV &6         &\bf{17.02}         &\textbf{78.76} \\
\midrule
\ha RVQuant~\cite{park2018value} &\cm&\cm &8&8      &22.47                  &74.22          \\
\hc \OURS       &\xm&\xm            &8 &8           &{22.47}         &\textbf{78.81} \\
\bottomrule 
\end{tabular}
}

\subfloat[\footnotesize SqueezeNext]{
\centering
\centering
\small
\setlength\tabcolsep{2pt}
\begin{tabular}{p{8em}ccccccccccccccccccccccccccccc} \toprule
    Method      &{No D}&{No FT} &W-bit&A-bit        &Size (MB)          &Top-1          \\
    \midrule
\ha Baseline    &--&--              &32&32          &9.86               &69.38          \\
\midrule    
\hc \OURS       &\xm&\xm            &\AV &8         &{1.23}      &\textbf{59.23} \\
\midrule    
\hc \OURS       &\xm&\xm            &\AV &6         &{1.85}      &\textbf{68.17} \\
\midrule
\hc \OURS       &\xm&\xm            &8 &8           &{2.47}      &\textbf{69.17} \\
\bottomrule 
\end{tabular}
}
\end{table*}

\begin{table}[!htbp]
\caption{Ablation study for \rg on ResNet18, ShuffleNet and SqueezeNext. 
We show the performance of \OURS with different sources of data to compute the sensitivity and determine the activation range.
All quantized models have the same size as quantized models with 4-bit weights.}
\label{tab:Abalation_SynD_extra}
\centering
\small
\setlength\tabcolsep{5pt}
\begin{tabular}{lcccccccccccccccccccccccccccccccccccccccccc} \toprule
Method                  & W-bit & A-bit     &ResNet18       &ShuffleNet    &SqueezeNext\\
\midrule        
\ha Baseline            &32&32              &71.47          &65.07  &69.38\\ 
\midrule        
\ha Gaussian            &\AV&8              &67.87          &56.23  &48.41\\
\ha Training Data       &\AV&8              &68.61          &58.90  &62.55\\
\midrule
\hc \rg               &\AV&8              &\textbf{68.45} &\textbf{57.50}&\textbf{59.23}\\
\bottomrule 
\end{tabular}
\end{table}

\subsection{Clipping}
\label{sec:clipping}
Quantization maps a single-precision tensor $z$ to a low-precision tensor $Q(z)$. This includes two steps: 1) clipping the original tensor to range $[a,~b]$, and then 2) mapping this range to integer range $[0,~2^{k}-1]$.
A simple way is to set $[a,~b]= [\min(z),~\max(z)]$ for conventional quantization methods. 
Recently, more effort has been spent on choosing the ``optimal'' range of $[a,~b]$~\cite{Kravchik_2019_ICCV, choi2018pact, banner2018post, li2019fully}, which are the so-called clipping methods.

In all of our experiments above, we use the simplest way, i.e., $[a,~b]= [\min(z),~\max(z)]$, to conduct the quantization. 
The main reason behind this is two-fold: 
(i) we want to show the efficacy of \OURS without the assistance of any other technique; 
(ii) some of proposed methods~\cite{banner2018post, choi2018pact} need hyper-parameter tuning to get the optimal $a$ and $b$ which can be costly.
However, we show that performance of \OURS can be further boosted by the weight clipping method, if the slightly higher  computational overhead could be afforded.
In particular, we use the ``percentile'' method proposed in~\cite{li2019fully}. 
This method directly clips a single-precision weight tensor to $\gamma$-th and $(1-\gamma)$-th percentiles (we refer the reader to~\cite{li2019fully} for more details). 
As shown in \tref{tab:imagenet_main_text} and \tref{tab:resnet18_dfc}, \OURS can be further improved by weight clipping.
